\documentclass[conference]{IEEEtran}
\IEEEoverridecommandlockouts
\usepackage{cite}
\usepackage{amsmath,amssymb,amsfonts}
\usepackage{algorithmic}
\usepackage{graphicx}
\usepackage{textcomp}
\usepackage{xcolor}
\usepackage{multicol}
\usepackage{graphicx} 
\usepackage{algorithm}
\usepackage{algorithmic}
\usepackage{float}

\usepackage{booktabs}
\def\BibTeX{{\rm B\kern-.05em{\sc i\kern-.025em b}\kern-.08em
    T\kern-.1667em\lower.7ex\hbox{E}\kern-.125emX}}

\title{AIoT-Based Drum Transcription Robot using Convolutional Neural Networks}

\makeatletter
\newcommand{\linebreakand}{%
  \end{@IEEEauthorhalign}
  \hfill\mbox{}\par
  \mbox{}\hfill\begin{@IEEEauthorhalign}
}
\makeatother

\author{
  \IEEEauthorblockN{Yukun Su}
  \IEEEauthorblockA{
    \textit{South China University of Technology}
    }
  \and
  
  \IEEEauthorblockN{Yang Yi}
  \IEEEauthorblockA{
    \textit{South China University of Technology}
    }
   
}

\begin{document}
\maketitle

\begin{abstract}
With the development of information technology, robot technology has made great progress in various fields. These new technologies enable robots to be used in industry, agriculture, education and other aspects.
In this paper, we propose a drum robot that can automatically complete  music transcription in real-time, which is based on AIoT and fog computing technology. Specifically, this drum robot system consists of a cloud node for data storage, edge nodes for real-time computing, and data-oriented execution application nodes. In order to analyze drumming music and realize drum transcription, we further propose a light-weight convolutional neural network model to classify drums, which can be more effectively deployed in terminal devices for fast edge calculations. The experimental results show that the proposed system can achieve more competitive performance and enjoy a variety of smart applications and services.

\end{abstract}


\section{Introduction}
The Artificial Intelligence of Things (AIoT) system~\cite{jazdi2014cyber,wan2011advances} is a multi-dimensional, complex system that integrates computing, networking, and physical environments. AIoT data is exchanged between different sensors, terminal cloud devices and edge/fog devices, enhancing the professionalism and pertinence of intelligent services through distributed computing design and network architecture. In recent years, with the continuous advancement of human society, robot technology has achieved unprecedented development, and the scope of its application fields is further expanded, which is manifested in the wider range of areas such as traditional industries, manufacturing, people's livelihood and so on.

As a component of AIoT, drum robot~\cite{crick2006synchronization,sui2020intelligent} is a hot topic in recent years, where the core problem is drum transcription. The goal of drum transcription is to generate drum events from audio signals~\cite{wu2018review}. While this is the first step for full analysis and fetching more music information of drum track, which plays an important role in automated drum education and entertainment systems.

In some previous relevant works~\cite{kotosaka2001synchronized,crick2006synchronization,ince2015towards}, they are more likely dependent on the preset sequence to play drums, which fail to analyze the online music and are lack of widespread use.
In addition, although some approaches propose the algorithms to extract and conduct the music transcription~\cite{gouyon2000use,herrera2002automatic,sandvold2004percussion}, the models they use are usually computational-costly and are hard to deploy in the smart devices like the chips. 
To this end, it is essential to combine AI~\cite{russell2002artificial} and IoT~\cite{lin2017survey,yaqoob2017internet} devices to improve the effectiveness and efficiency of drum transcription robot. Among them, AI algorithms can save human resources and improve efficiency, achieving higher accuracy and precision. At the same time, AI helps drumming robots to learn common rules and features of music from humans. The IoT enables objects to be sensed and controlled remotely across existing network infrastructure, including the Internet, thereby creating opportunities for more direct integration of the physical world into the cyber world. The IoT becomes an instance of cyber physical systems with the incorporation of sensors and actuators in IoT devices. Various IoT clusters generate huge amounts of data from diverse locations, which creates the need to process these data more efficiently.

In this paper, we introduce an AIoT-based drum transcription drum robot, which consists of a cloud node for data storage, edge nodes for real-time computing, and data-oriented execution application nodes. The cloud stores the required audio sequences and network files which attempts to schedule computationally demanding tasks with low communication requirements due to its abundant computing resources. To meet real-time performance in the real world, we use fog computing method, through which the communication intensive tasks with low computational demands can be processed in the fog at a fast speed without excessive reliance on cloud nodes. 

In order to employ our drum transcription algorithm into the robot, we further propose a light-weight convolutional neural network for computing. By using the AI chip and embedded board with neural computation acceleration stick, we can deal with some methods in real-time on the fog node and enable the drumming robot to process the tasks of drum extraction and classification.

Our main contributions are as follows:
\begin{itemize}
	\item [1)]
	We propose an AIot-based drum transcription robot prototype, which consists of a cloud node for data storage, edge nodes for real-time computing, and data-oriented execution application nodes.
	\item [2)]
	To meet the demand of real-time computing, we further propose a light-weight convolutional neural netwrok. Compared to the standard convolution operation, we can reduce the network parameters and network complexity.
	\item [3)]
	The experimental results show that the proposed system can achieve more competitive performance.
\end{itemize}

\begin{figure}
	\begin{center}
		\centering
		\includegraphics[width=3.3in]{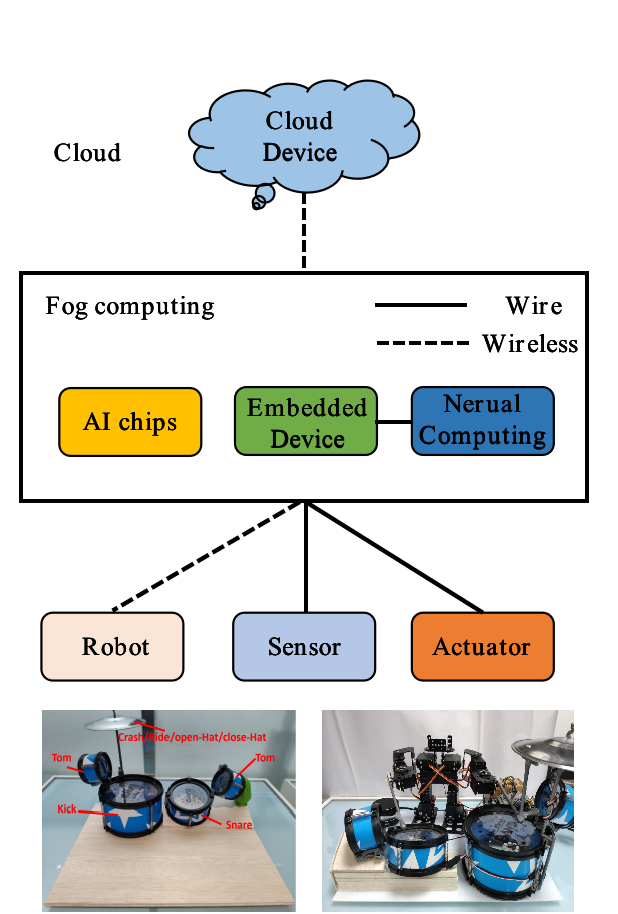}
	\end{center}
	\caption{The overview system of the drum robot.}
	\label{fig1}
\end{figure}

\section{Related Work}

\subsection{AIoT Computing}

Fog computing and related edge calculations have become a system model for reducing network congestion and latency for fog devices and applications, offloading some of the data originally located in the cloud to fog computers, thus improving real-time performance. The combination of every available resource (Cloud, Clusters, low-cost devices, and Smartphones)  can result in a significant decrease in the execution time and expenses~\cite{dimitrios2018simulation,zhu2019fog}. In~\cite{stavrinides2019hybrid}, Stavrinides et al. propose a hybrid approach to scheduling real-time IoT workflows in fog and cloud environments in a three-tiered architecture.
Currently, there is no common paradigm for fog computing. However, this idea has led to many domain-specific computing architectures. ~\cite{nasir2019fog} proposes a framework for a distributed summarization of surveillance videos over the fog network. The network itself is composed of multiple regions combined as clusters of Raspberry Pi. ~\cite{hsu2018creative} presents a creative IoT agriculture platform for cloud fog computing. The use of multiple Raspberry Pi devices creates a simulated decentralized environment. The platform allows farmland with limited network information resources to be integrated and automated, including agricultural monitoring automation, pest management image analysis, and monitoring.

\subsection{Drum Robot}

Sui~\cite{sui2020intelligent} proposed an intelligent human interaction robot by using SVM classifier~\cite{joachims1998making}. In~\cite{kotosaka2001synchronized}, Kotosaka presented a network of nonlinear oscillators for their robot system to synchronize with the human drummer. Crick et al.~\cite{crick2006synchronization} also designed a multi-sensor data fusion method, including visual and auditory data, which enables a robot to drum in synchrony with human performers. In another study, Ince et al.~\cite{ince2015towards} presented a framework for drum stroke detection and recognition by using auditory cues. Based on turn-taking and imitation principles, they designed an interactive drumming game, in which the participants improved their ability to imitate by using the proposed framework.
Using a remote server with powerful computing resources to process the machine  and learning method and then deploy it back to the robot is the approach taken by most cloud-based computing robots~\cite{liu2019novel,zheng2018cognition} nowadays. However, the audio processing on a remote server takes longer to consume in the transmission of the network than fog computing.

\begin{figure*}
	\begin{center}
		\centering
		\includegraphics[width=7.0in]{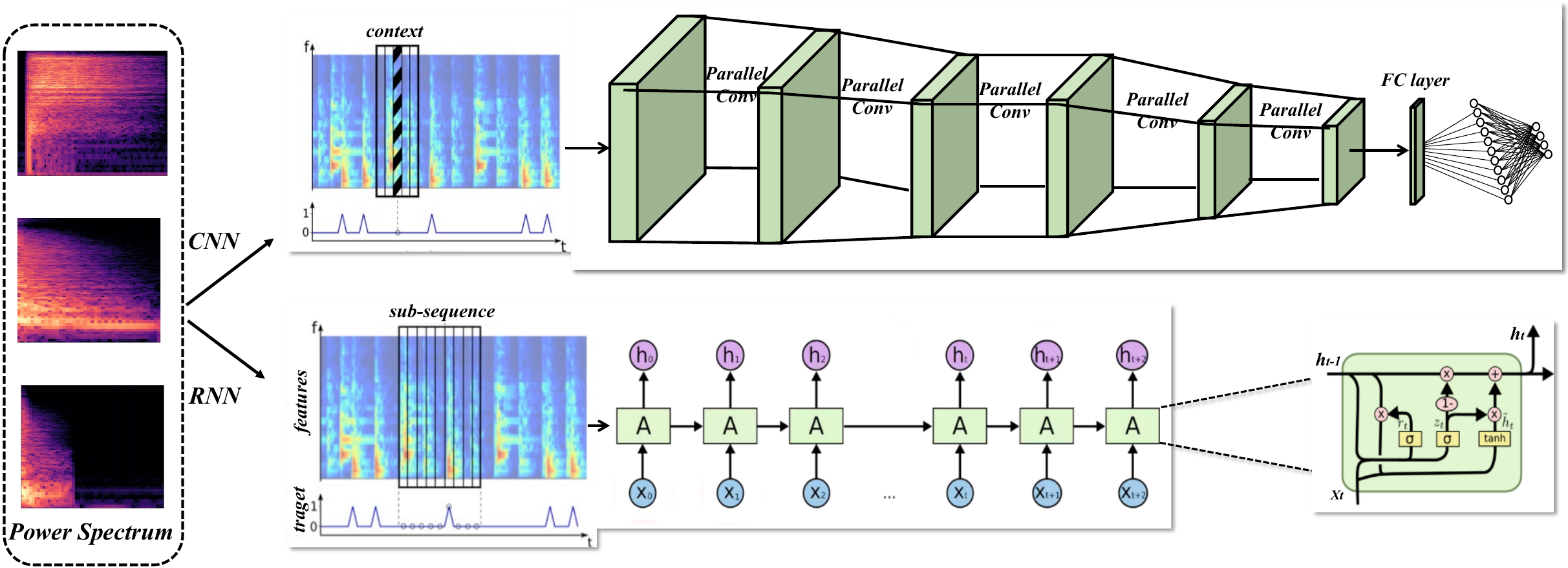}
	\end{center}
	\caption{An overview of drum transcription. We first extract the audio and convert it into the power-spectrogram, then the network learns from the 2D image information. Specifically, we here compared two different deep neural networks: CNN and RNN.}
	\label{fig2}
\end{figure*}

\subsection{Deep Learning Network}
Deep neural networks are widely used in classification~\cite{krizhevsky2012imagenet,su2020human}, detection~\cite{su2022epnet,su2022self} and segmentation~\cite{su2021context,su2023unified}
However, deep neural network acceleration is moving toward smaller~\cite{li2019light}.
Unlike other methods of model compression and acceleration, compact networks can be trained from the scratch without the need for fine-tuning and retraining to achieve convergence. In addition, other methods of model compression can be continued on the basis of lightweight networks, which can be compressed and accelerated at a deeper level. In view of these advantages, various lightweight network architectures are proposed.
The increasing need for running high-quality deep neural networks on limited computational devices is the heat trend on efficient model designs~\cite{design}. SqueezeNet~\cite{squeezenet} reduces parameters and computation significantly using 1$\times$1 and 3$\times$3 convolutional filter concating while maintaining accuracy. Compared to merely increasing the depth and width of the network, GoogLeNet~\cite{googlenet} introduces a module with dimension reductions within lower complexity. SENet~\cite{senet} utilizes the efficient block to achieve impressive performance at a slight computation cost. 
Group convolution was first introduced in AlexNet~\cite{alexnet} for distributing the model over two GPUs. Later it has been well revealed its effectiveness in ResNeXt~\cite{resnext}. As a special kind of group convolution, depthwise separable convolution was initially introduced in ~\cite{DWconv} and subsequently used in Inception~\cite{inceptionv3,incpetinv4}. Currently, MobileNet~\cite{mobilenet,mobilenetV2} and ShuffleNet~\cite{shufflenet,shufflenetV2} utilizes these convolutional methods and gains state-of-the-art results among lightweight models.

\section{Drumming Robot System }
The proposed AIoT-based drum transcription robot system is shown in Figure \ref{fig1}. The bottom layer of the framework is the device layer, which contains hardware devices such as robots, sensors, and actuators. The middle layer is a fog layer consisting of a distributed embedded board and a neural computing stick as well as an AI chip. These three types of devices form the fog computing platform. We use multiple embedded boards and AI chips as nodes for fog calculation to implement a distributed computing architecture. At the same time, these intelligent embedded boards on the fog side can also provide some computing resources as decision nodes of the system. Embedded boards must have high computational performance and low power consumption, which is responsible for sensor information processing and actuator control. The system uses deep neural network algorithms for drum transcription. Current algorithms consume a lot of computation power. As a result, the we propose a light-weight neural convolutional network that can accelerate deep network reasoning with lower computation power consumption. The top layer of the framework is the cloud side, which consists of large databases that store data such as models and music.

\section{Methodology}

\subsection{Audio Extraction}
First, the input audio is converted into a spectrogram using short-time Fourier transform (STFT). The STFT is calculated with librosa~\cite{mcfee2019librosa} using a Hanning window with 2048 samples window size and 512 samples hop length. Then, the spectrogram is computed using a Mel-filter~\cite{farooq2001mel} in a frequency range of 20 to 20000 Hz with 128 Mel bands, resulting in a 128 $\times$ $n$ power-spectrogram.

\subsection{Overall Architecture}
As shown in Fig~\ref{fig2}, the overall architecture of our proposed method takes the converted 2D power-spectrogram audio data as input. Then, we encode the data by leveraging the deep neural learning networks such as CNN~\cite{krizhevsky2012imagenet} or RNN~\cite{lipton2015critical}.
Note that in our method, we use CNN as our network since the input data is in 2D image form, and the designed light-weight CNN model is more widely used. For comparison, we also depict the RNN in Fig~\ref{fig2}.

\subsection{Light-weight Convolution}
A standard convolutional~\cite{standardconv} layer takes as input a ${\mathit D_H}$ $\times$ ${\mathit D_W}$ $\times$ ${\mathit D_M}$ feature map \textit{F} and produces a ${\mathit D_H}$ $\times$ ${\mathit D_W}$ $\times$ ${\mathit D_N}$ feature map where ${\mathit D_H}$ and ${\mathit D_W}$ is the spatial height and width of a square input feature map, ${\mathit D_M}$ is the number of input channels (input depth), ${\mathit D_N}$ is the number of output channel (output depth), $\mathit D_K$ is the kernel size. 

\begin{figure*}
	\begin{center}
		\centering
		\includegraphics[width=6.8in]{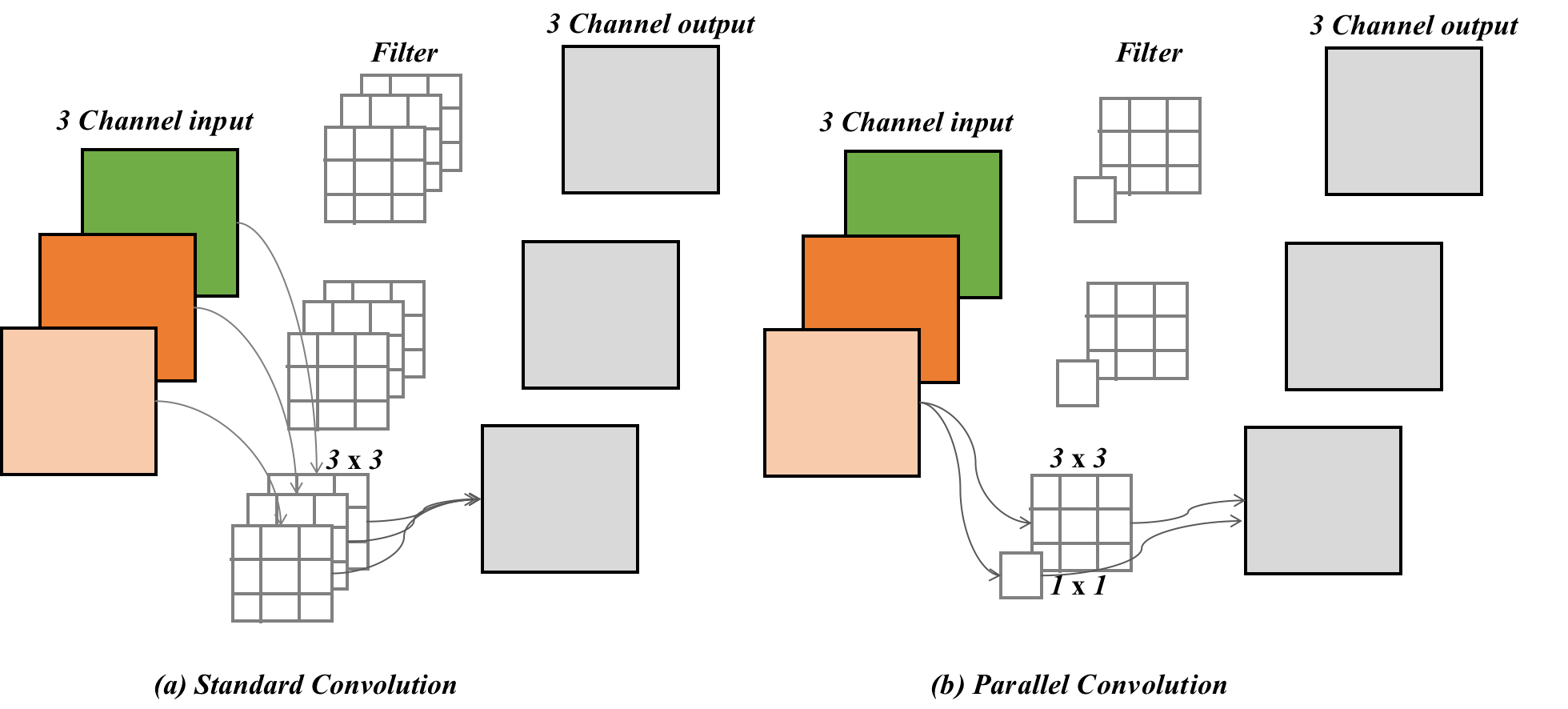}
	\end{center}
	\caption{Different convolution operation structure diagram. (a) Standard convolution uses all three input channels with all the 3$\times$3 filters for operation, which is computational-costly. (b) Our proposed Parallel convolution simply replaces the 3$\times$3 filters with one 3$\times$3 and one 1$\times$1 filters, and then these two filters are responsible for each input channel.}
	\label{fig3}
\end{figure*}

Therefore the total computational cost at this layer $L$ can be given as:
\begin{equation}
F_{L_{Standard}}=\mathit D_H \cdot \mathit D_W \cdot \mathit D_K \cdot \mathit D_K \cdot \mathit D_M \cdot \mathit D_N
\label{eq1}
\end{equation}

Common CNNs adopt a spatially layered architecture. With the spatial dimensions of the next feature map reduced by downsampling, the feature information can be extracted to avoid over-fitting while reducing the amount of calculation. Due to limited computing resources, compact networks are affected by both weak feature representations and limited information capacity. Different downsampling strategies give a compromise between the detailed feature representation of the CNNs and the large information capacity. Slow downsampling strategy performs in later layers of the network, so more layers have large spatial dimensions. Instead, downsampling is performed at the beginning of the network in a fast downsampling strategy, which significantly reduces computational costs. Thus, given a fixed computational budget, slow downsampling strategies tend to generate more detailed features, while fast downsampling strategies can increase the number of channels and allow more information to be encoded. When computing resources are extremely limited, information capacity plays a more important role in network performance. Traditionally, the number of channels can be reduced to accommodate a compact network architecture to some complexity. With a slow downsampling scheme, the network becomes too narrow to encode enough information, which results in severe performance degradation. 

Noticing Eq~\ref{eq1} and it can be clearly seen that the computational cost depends on the kernel size (${\mathit D_K}$), the feature map size (${\mathit D_H}$ and ${\mathit D_W}$), the input channel ${\mathit D_M}$ and the output channel ${\mathit D_N}$. In general, the size of the filter ${\mathit D_K}$ of the convolutional neural network is 3$\times$3, and changing the number of channels of the feature input and output arbitrarily will change the representation ability of the network. Therefore here we take a fast downsampling approach. By downsampling at the previous network's layers, the input feature map size (${\mathit D_H}$ and ${\mathit D_W}$) is reduced, which results in a decrease in the amount of calculation.

\begin{table}[]
	\begin{center}\caption{The Light-weight CNN (Parallel Convolution Network) Architecture.}
	\label{table1}
		\scalebox{1.0}{
			\begin{tabular}{cc}
				\toprule
				Light-weight CNN Architecture & Output Channel\\
				\midrule
				Parallel-Conv &64\\
				Parallel-Conv &64\\
				Down Sampling (AvgPooling)\\
				Parallel-Conv &128\\
				Parallel-Conv &128\\
				Down Sampling (AvgPooling)\\
				Parallel-Conv &256\\
				Parallel-Conv &256\\
				Parallel-Conv &256\\
				Down Sampling (AvgPooling)\\
				Parallel-Conv &512\\
				Parallel-Conv &512\\
				Parallel-Conv &512\\
				Parallel-Conv &512\\
				Parallel-Conv &512\\
				Down Sampling (AvgPooling)\\
				Parallel-Conv &1024\\
				Parallel-Conv &1024\\
				FC-num\_classes\\
				\bottomrule
		\end{tabular}}
	\end{center}
	
\end{table}

To this end, in this paper, we propose a \textbf{Parallel Conv} operation in our network and replace the standard convolution in each layer in the network to reduce the model parameters.
As shown in the Fig-\ref{fig3}, compared to the traditional standard convolution, which performs convolution using all the input channels with all the 3$\times$3 filters, by contrast, the convolution kernels in parallel convolution execute 3$\times$3 convolution and 1$\times$1 convolution at the same time, which can be regarded as a combination product of 3$\times$3 group convolution and 1$\times$1 point-wise convolution. 
Formally, for each input channel, we make it undergo one 1$\times$1 and  one 3$\times$3 and then yield the output channel. The main purpose of 1$\times$1 kernel is to apply nonlinearity. After each layer of the neural network, we can apply an activation layer. To increase the number of nonlinear layers without significantly increasing the parameters and computation, we can apply a 1×1 kernel and add an activation layer after it, which can help to add more layers of depth to the network. Table~\ref{table1} also shows us the network architecture.

\begin{table}[]
	\begin{center}\caption{\textbf{Table shows the results for different algorithms on drum transcription task.}}
	\label{table2}
		\begin{tabular}{cc}
			\hline
			Method & Accuracy(Top1\%)\\
			\hline\hline
			Sui $\emph{et~al.}$~\cite{sui2020intelligent} SVM &82.50 \\
			RNN &91.87 \\
			\textbf{light-weight CNN} &\textbf{92.18}\\
			\hline
		\end{tabular}
	\end{center}
\end{table}

\begin{table}[]
	\begin{center}\caption{\textbf{The comparison of the number of parameters and computational complexity.}}
	\label{table3}
		\begin{tabular}{ccccc}
			\hline
			Method & Parameters (M)& GFLOPs & Accuracy(Top1\%)\\
			\hline\hline
			CNN & 27.16 & 32.54& 93.4\\
			\textbf{light-weight CNN} &\textbf{20.54} &\textbf{23.82} &\textbf{92.18} \\
			\hline
		\end{tabular}
	\end{center}
\end{table}

More specifically, in the case of parallel convolution with the group size of ${\mathit g}$, the total computational cost for layer $L$ can be computed as:
\begin{equation}
F_{L_{Parallel}}=\mathit D_H \cdot \mathit D_W \cdot \mathit D_K \cdot \mathit D_K \cdot \frac{\mathit D_M}{g} \cdot \mathit D_N + \mathit D_H \cdot \mathit D_W \cdot \mathit D_M \cdot \mathit D_N
\label{eq2}
\end{equation}

Therefore the total reduction (${\mathit R}$) in the computation as compared to the standard convolution:
\begin{equation}
\begin{split}
\mathit R_{Parallel/Standard} &=\frac{F_{L_{Parallel}}}{F_{L_{Standard}}}  \\ \notag
& = \frac{1}{\mathit g} + \frac{1}{\mathit K^2} < 1
\end{split} \tag{3}
\label{eq3}
\end{equation}

\section{Experiments}
\textbf{Datasets:}  For the files of the drums audio, we collected various categories of sounds (Tom, Kick, Snare, Close-Hat, Ride, Crash and Open-hat) in the various open-source databases~\cite{dittmar2014real,gillet2006enst} on the Internet. Then we divided them into the training sets and verification sets.

\textbf{Performance:} For the comparison of the results, we conducted experiments on the verification set. The experimental results are shown in Table~\ref{table2}. The results show that CNN works best. This shows that CNN has been successfully applied not only in image processing but also in many other deep learning tasks. Though CNN has no memory function to process time series, the spectral context allows CNN to access surrounding information in the same way that it processes images during training and classification. In the meantime, CNN can provide translation invariance convolution in time and space, as well as the advantage that pitches invariant kernels can be easily learned by CNN, so convolution can be used to overcome the diversity of audio signal and it is well-equipped to learn an acoustic model task. In this work, the input will be treated as a specific image, then the convolutional layer, max-pooling layer, and drop-out layer finish the task of extracting and learning feature maps.

\begin{table}[]
	\begin{center}\caption{\textbf{Performance of speed of processing audio inference tasks between cloud and fog node.}}
	\label{table4}
		\begin{tabular}{c|cccc}
			\toprule
			Node& Inference device& T-time& C-Time& Total Time\\
			\midrule
			Cloud& CPU& 560 $\pm$ 20 ms& 14.37 ms& 574.37 $\pm$ 20 ms\\
			Cloud& GPU& 560 $\pm$ 20 ms& 9.62 ms& 569.62 $\pm$ 20 ms\\
			Fog& Pi+NPU& —& 13.36 ms& 13.36ms\\
			Fog& RK3399pro& —& 11.68 ms& 11.68ms\\
			\bottomrule
		\end{tabular}
	\end{center}
\end{table}

\begin{figure}
	\begin{center}
		\centering
		\includegraphics[width=3.3in]{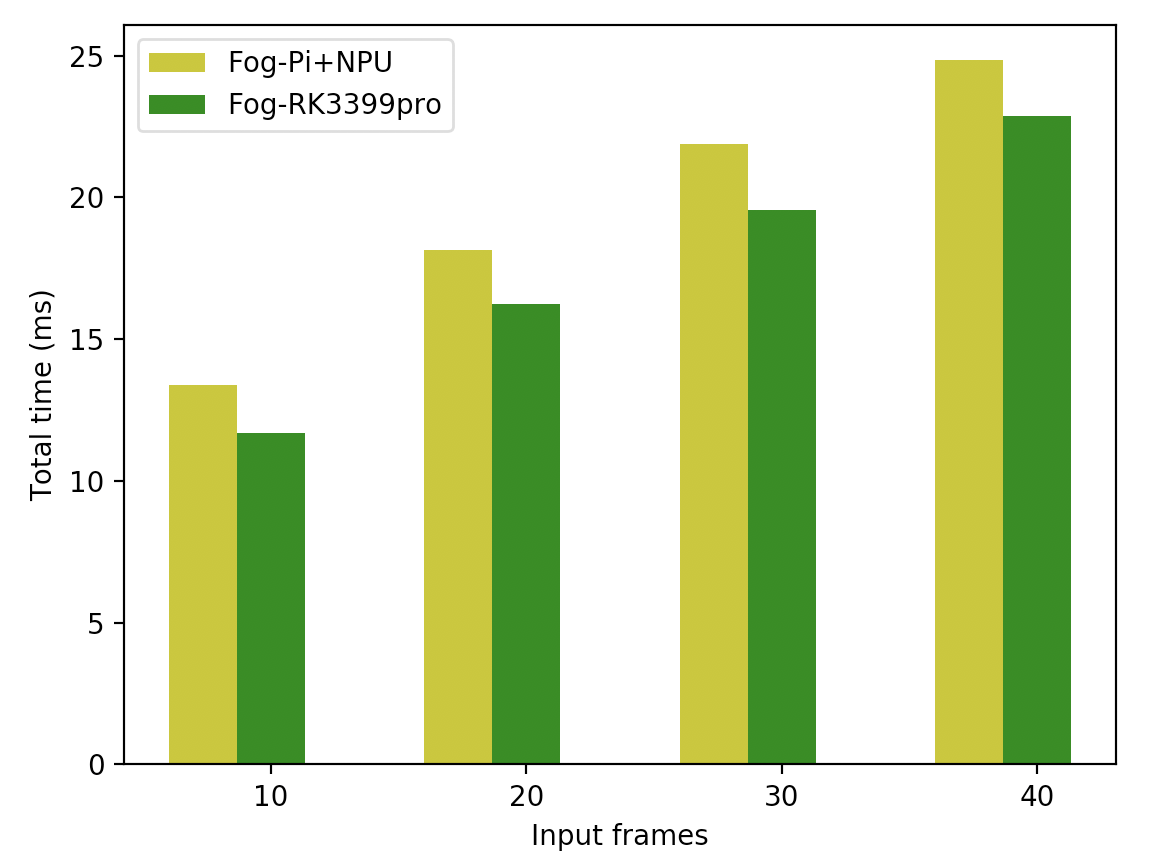}
	\end{center}
	\caption{Histogram of performance with different input frames.}
	\label{fig4}
\end{figure}

We also compare the performance between the standard CNN and our proposed light-weight CNN. As shown in Table~\ref{table3}, our method can achieve competitive performance while reducing a large amount of network parameters and complexity.

Furthermore, in the fog computing model, we also conduct a number of experiments comparing the overall speed of the same task between the two computing nodes on the cloud side and the fog side. 
In the cloud, we use CPU and GPU in the PC separately. 
On the fog side, we use Raspberry Pi equipped with neural processing unit (NPU) to accelerate the calculation and RK3399pro AI chip which integrated neural network processor NPU.
Its computing power can be up to 3.0Tops, compatible with a variety of AI frameworks, and it can be combined with the backplane to form a complete industrial application board. 
At the same time, we define the number of frames of the input file to be consistent (10 frames), and define T-time as the transmission time, and C-time as the computing time. The comparison results are shown in Table~\ref{table4}. 
It can be seen that the calculation at the cloud side can achieve the fastest speed on the GPU, but the delay of the network transmission has a negative impact on the total processing time. In the calculation of the fog side, by using the NPU accelerator stick on the Raspberry Pi and the AI chip, their processing speed is no less than that of the CPU on the PC, and they have no delay in the network transmission, and the final total processing speed can reach real-time. 
The results in Fig~\ref{fig4} show the processing time of tasks with different input frames on different computing devices at different fog nodes.

All the above results show that computing in the fog node is better than the other two methods, and the real-time processing effect can be achieved.

\section{Conclusion}

In this paper, we present a novel drum transcription robot using fog computing, which consists of a cloud node for data storage, edge nodes for computing, and data-oriented execution application nodes. 
To further meet the demands of real-time computing, we propose a light-weight convolutional neural network to reduce the network parameters and complexity. 
In our future work, more resources on smartphone and other mobile devices will be more considered and utilized, and the algorithms for audio processing needs to be further improved.

\bibliographystyle{IEEEtran}
\bibliography{ref}

\end{document}